\documentclass[journal,onecolumn]{IEEEtran}

\usepackage[utf8]{inputenc}
\usepackage{multirow}
\usepackage{graphicx}
\usepackage{indentfirst}
\usepackage[small]{caption}
\usepackage[ruled]{algorithm2e}
\usepackage{multirow}
\usepackage{booktabs}

\ifCLASSINFOpdf
\else
\fi

\begin{document}
%
\title{A self-adapting super-resolution structures framework for automatic design of GAN}
%
%
%

\author{Yibo~Guo,
        Haidi~Wang, Yiming~Fan, shunyao ~Li,
        and~Mingliang ~Xu,
\thanks{Y. Guo is a teacher at the School of Information Engineering, Zhengzhou University}
\thanks{H. Wang, Y. Fan, S. Li and M. Xu are with Zhengzhou University.}
}

%
%

\markboth{Journal of \LaTeX\ Class Files,~Vol.~14, No.~8, August~2015}%
{Shell \MakeLowercase{\textit{et al.}}: Bare Demo of IEEEtran.cls for IEEE Journals}
%



\maketitle

\begin{abstract}
With the development of deep learning, the single super-resolution image reconstruction network models are becoming more and more complex. Small changes in hyperparameters of the models have a greater impact on model performance. In the existing works, experts have gradually explored a set of optimal model parameters based on empirical values or performing brute-force search. In this paper, we introduce a new super-resolution image reconstruction generative adversarial network framework, and a Bayesian optimization method used to optimizing the hyperparameters of the generator and discriminator. The generator is made by self-calibrated convolution, and discriminator is made by convolution lays. We have defined the hyperparameters such as the number of network layers and the number of neurons. Our method adopts Bayesian optimization as a optimization policy of GAN in our model. Not only can find the optimal hyperparameter solution automatically, but also can construct a super-resolution image reconstruction network, reducing the manual workload. Experiments show that Bayesian optimization can search the optimal solution earlier than the other two optimization algorithms.
\end{abstract}

\begin{IEEEkeywords}
Single image super-resolution, Generative adversarial network, Self-calibrated convolutions, Bayesian optimization.
\end{IEEEkeywords}

\IEEEpeerreviewmaketitle

\section{Introduction}
%
%
%
%
Super-resolution images have rich detailed information, which can effectively assist in completing tasks such as medical image analysis\cite{Lally2020Unbalanced} and object recognition\cite{Bai2018Finding}. Super-resolution image reconstruction aims to use the neural network to learn the mapping relationship between low-resolution images and high-resolution images. With the development of deep learning, the network structure has become more and more complex, and the model parameters have gradually increased. Model parameters include the network random initialization parameters and hyperparameters that can be manually adjusted. The purpose of neural network training is to obtain a set of parameters to make the model have better generalization capabilities. Small changes in hyperparameters may have a greater impact on the generalization of the model. A model has only one set of parameters to make the model optimal. In order to obtain the optimal performance of the model, it is necessary to gradually research the network model hyperparameters when the network model framework has been fixed. In traditional method, Deep learning model usually adopts manual tuning of model hyperparameters, and gradually explores the optimal solution with reference to experience values. This process takes a long time and easily makes people impatience. An automated method to adjust parameters is easier to be accepted by researchers, and an automated exploration of model parameters can advance deep learning research. Therefore, we try to build a generative adversarial network model with optimization algorithms and reduce the manual workload as much as possible.

Grid search is a traditional method of neural network hyperparameter search, which can realize neural network auto-tuning. In the process of parameter search, grid search will list all combinations of parameters. For deep learning networks, the computation is relatively large, and it requires more processor resources. Some scholars have proved that the random search algorithm \cite{Florea2019Weighted} is superior to the grid search algorithm, but the random search algorithm has strong randomness which is not conducive to the optimization of neural network.

Bayesian optimization algorithm \cite{Pelikan1999BOA} is a black-box optimizers (that assume no internal analytical structures of the objective function), which supports the above-mentioned deep learning network training. In addition, the Bayesian optimization algorithm is to predict the next evaluation point based on the historical results, which can solve the problem of large computation and resources for grid search and random search. In this paper, the Bayesian optimization method is used for deep learning optimization to achieve the construction of generative adversarial networks.

In our work, we propose generative adversarial network model with the Bayesian optimization algorithm. It can search for a set of optimal model hyperparameters and reduce manual workload. Our second contribution is using the constructed generative adversarial network model for super-resolution image reconstruction tasks. We compare the reconstructed image results of the three optimization algorithms under different public datasets, and the results of Bayesian optimization are the best.

The rest of our work is organized as follows: Section 2 introduces the research of super-resolution reconstruction methods and several different algorithms of hyperparameter search. We have a brief discussion of the advantages of GAN method and Bayesian optimizations. Section 3 builds a super-resolution image framework based on the generative adversarial networks and determines GANs hyperparameters through Bayesian optimization algorithm. In section 4, we compare the speed of three optimization algorithms and the quality of reconstructed images used diffenert algorithms, which illustrates the effectiveness of proposed methods. In section 5, there are the conclusion of this work and the future perspectives.

\section{Related Work}

This chapter mainly introduces the current research status in two aspects, one is the current research status of super-resolution image reconstruction based on generative adversarial networks, and the other is the current research status of optimization algorithm.

1) Super-resolution image reconstruction method based on generative adversarial networks

The Generative Adversarial Network (GAN) \cite{Goodfellow2014Generative} is consisted of two networks: a generator and a discriminator. The two networks were trained through a max-min game and could generate more realistic virtual samples. Generative adversarial networks were more sensitive to image colors and brightness. The use of generative adversarial networks for super-resolution reconstruction had better visual effects, and some progress had been made. The idea of image inpainting \cite{Liu2018Image} was referenced from image outpainting \cite{Sabini2018Painting}. Extending the center of the image to both sides could generate a high-resolution image, but the generated extended area was blurred, and the boundary with the center of the image was clearly visible. Pix2pixHD \cite{Wang2017High} used multiple stages to gradually obtain super-resolution images, and the discriminator was also modified to multi-scale discriminating the input image. CRRAGAN \cite{Chen2021Cascading} proposed a cascaded residual-residual attention extraction module, which used feature fusion for information forward propagation, and achieved good results.Most of the above methods used multi-scale or cascaded residual methods to extract image features. The use of extracted features through feature fusion had become a common way to build network structures and was widely used.  Chudasama \emph{et al.} \cite{Chudasama2021Computationally,Chudasama2021RSRGAN} proposed two different effective calculation methods for single image super-resolution, which were reduced the number of parameters on the basis of achieving better results. Reducing the model parameters could reduce the tuning time of model hyperparameter to a certain extent, but some hyperparameters were essential. The method in this paper adopted another method to save the tuning time, which was automatically searched for a set of model hyperparameter combinations through optimization algorithms.The following introduces some universal model hyperparameter search algorithms.

2) Optimization algorithm

In the deep learning model of super-resolution reconstruction, small changes in hyper-parameters might have a greater impact on the performance of the model. An automated hyperparameter tuning algorithm is beneficial to the automatic construction of the model. Bergstra \emph{et al.} \cite{Bergstra2011Algorithms} explored several machine learning hyperparameter optimization algorithms, including random search and two greedy algorithms. At the same time, they proved that random search \cite{Florea2019Weighted} was better than grid search algorithm. In addition to the above optimization methods, the most famous algorithm is the Nelder-Mead(NM) \cite{Audet2017Nelder}, which was an unconstrained minimization method and was widely used \cite{Hz2020Orthogonal,Niegodajew2020Power}. The NM algorithm did not need to consider the gradient information of the function, which might cause a dimensional disaster problem. It is a black box optimization algorithm. COBYLA \cite{Powell1994A,Kramer2016Scikit} was inspired by the NM algorithm, and it was an improved nonlinear constrained optimization method. The objective function was approximated by linear interpolation and confidence interval constraint variables at the vertices of the simplex. After calculation, a new variable vector was obtained. Variables that met certain conditions would replace the current variables and got the current optimal vector. However, the linear approximation of this algorithm was inefficient, and variables could not be too many. Different from the above optimization algorithms, particle swarm optimization(PSO) \cite{Kennedy2002Particle} was a kind of evolutionary algorithm, which was more sensitive to dimensionality. PSO proposed a particle swarm optimization method for optimizing nonlinear functions and neural network training. If the dimensionality was too high, it would seriously increase the amount of function evaluations. Gong \emph{et al.} \cite{Gong2020Research} used gene expression programming to optimize hyperparameters. The evolutionary algorithm could search for the global optimal hyperparameter combination through fuzzy control technology. The evolutionary algorithm took longer to evaluate the function as dealing with the large-scale problems, and it might involve topological structure and geometric calculations.

Bayesian optimization \cite{Pelikan1999BOA} did not involve the above-mentioned topological structures and geometric calculations, and it also didn't analyze the structure of the objective function. It was a black-box optimization algorithm which selected the location of the evaluation point through the acquisition function. The acquisition function ensured that the next evaluation point was always optimal or explored points which had not appeared before. Bayesian optimization useed the covariance matrix to solve the above-mentioned dimensionality problems. Therefore, Bayesian optimization was often used for a variety of research tasks. For example, Liu \emph{et al.} \cite{Liu2021Industrial} proposed to use Bayesian optimization for intrusion detection tasks, searching for a set of optimal hyperparameters suitable for the network to improve the detection accuracy. And Xiang \emph{et al.} \cite{Xiang2020A} proposed the use of Bayesian optimization to construct the skeleton of robots and animations to extend the automatic construction method of the skeleton. Xu \emph{et al.} \cite{Xu2020A} used Bayesian optimization for generator parameter search and realized the generation of time series data. The music generated by this method could not be distinguished by humans. Our approach was different from the time series prediction model and the method for automatically building the skeletons. In this paper, it is mainly aimed at the construction of a generative adversarial networks model for image reconstruction, rather than the problem of sample generation. Inspired by the automatic skeleton construction method, we use Bayesian optimization algorithm to optimize model. An automatical method can reduce the manual workload and realize adaptive construction of network models.

\section{Method}

Automated design can speed up manual parameter adjustment. In this paper, a Bayesian optimization method is used to optimize the generative adversarial network. We realize automatic parameter adjustment and avoid the tedious process of manual parameter adjustment. Different from the method proposed by Xiang \emph{et al.} \cite{Xiang2020A}, which mainly adjust the parameters of the joint points connected by the robots and the animation skeletons. Our method is to tune the hyperparameters of the deep learning model and build the generative adversarial network structure automatically.

The optimization algorithm aims at achieving the minimum value of the optimization objective function, and gradually searches for the global optimal solution. In the neural network training process,  the objective function structure information cannot be obtained. In this paper, Bayesian optimization algorithm is used to optimize generative adversarial network model. Compared with other algorithms, Bayesian optimization has the following advantages: (1) It is a black box optimization method, without considering the objective function structure and gradient information; (2) No complicated topological structure is needed to assist in searching the optimal solution.

In this paper, we build a super-resolution image reconstruction model based on generative adversarial network by Bayesian optimization algorithm. It is mainly divided into two stages: on the one hand, we build a basic generative adversarial network model framework and define the parameters of generator and discriminator; on the other hand, it is the optimization process of network training combined with Bayesian optimization algorithm.

\subsection{Super-resolution image reconstruction model based on self-calibrated convolutional GAN}

In this paper, the method refers to the SRGAN \cite{Ledig2016Photo} model to construct the network structure, and the traditional method refers to empirical values to set the network structure and parameters. Different from the SRGAN parameter setting, we would use an automated method to explore the model parameters. Therefore, the residual module of the generator network in the SRGAN network is replaced with a self-calibrated convolution module. This module extracts features more adequately and facilitates to select super parameters by optimization algorithms. In order to realize the self-adaptive construction of the network structure, we parameterizes the components of the generator and the discriminator. Finally, we realize the self-adaptive construction of the generative adversarial network model through Bayesian optimization policy to adjusts the hyperparameters.

For deep learning model hyperparameter optimization tasks, the complexity of the network increases the difficulty of automatic parameter adjustment. There are many hyperparameters in the network model. Too many optimization parameters not only increase the optimization time, but also prone to NP-hard problem. The main problem of optimization is to control the number of hyperparameters. In addition, generative adversarial networks are prone to pattern collapse. Therefore, we adopts a fixed training strategy and only optimizes hyperparameters such as the number of network layers and the number of neurons. In this paper, the generator adopts a self-calibrated convolution module to extract low-resolution image features. The generator contains two hyperparameters, which are the number of self-calibrated convolution modules layers and the number of neurons in the self-calibrated convolution. We parameterize the generator, where m represents the number of self-calibrated convolutional modules layers, and n represents the number of neurons. m and n both are positive integers, and n must be a multiple of 4, namely $n\% 4 = 0$. The training of the generative adversarial network is a max-min game process. In order to avoid the problem of too many optimization parameters, and the discriminator has a simpler structure than the generator. Therefore, we only define one parameter in the discriminator. The discriminator encapsulates the convolutional layer, the batch normalization layer, and the activation function layer into one module. Exploring through the optimization model, the discriminator contains several layers of this module, where k represents the number of layers of the module contained in the discriminator, and k is a positive integer.

\begin{figure}[t]
\centering
\includegraphics[scale=0.6]{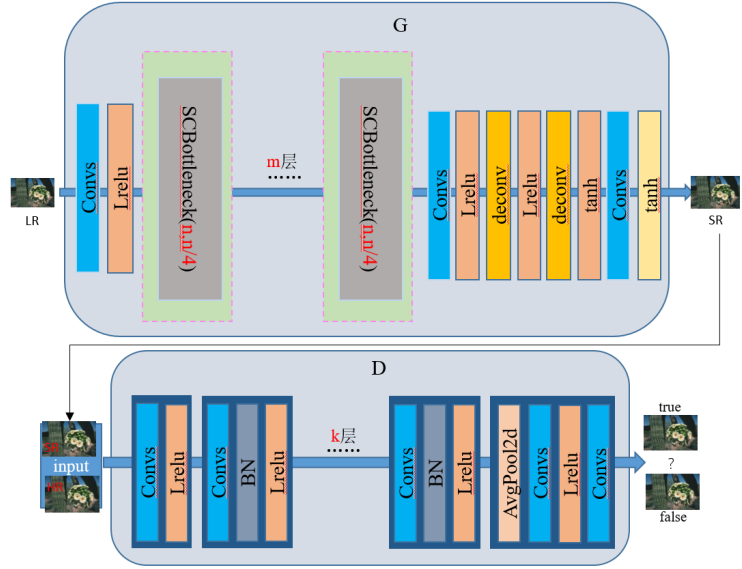}
\caption{The architecture of GAN model with parameters}
\label{fig:1}
\end{figure}

This chapter will encapsulate the structure of the generative adversarial network so that it can meet the hyperparameter tuning form and facilitate to achieve Bayesian optimization. The generative adversarial network model with parameters is shown in figure \ref{fig:1}.

\subsection{Bayesian optimization}

In deep learning, Bayesian optimization usually needs to consider four aspects to obtain the minimum value of the objective function through the network model: objective function, bound range, historical result record and optimization method. The explicit functions are as follows:
\begin{itemize}
\item Objective function: the objective to be optimized, the performance evaluation of the test set or verification set.
\item Bound range: selecting specific hyperparameter, and setting the upper and lower bounds of the hyperparameters. It defines the value range
\item Optimization method: it defines an acquisition function which determines the next evaluation point based on the posterior distribution of the optimization target.
\item Historical result record: recording the previous target value, as a reference for determining the next evaluation point.
\end{itemize}

\begin{figure}[t]
\centering
\includegraphics[scale=0.6]{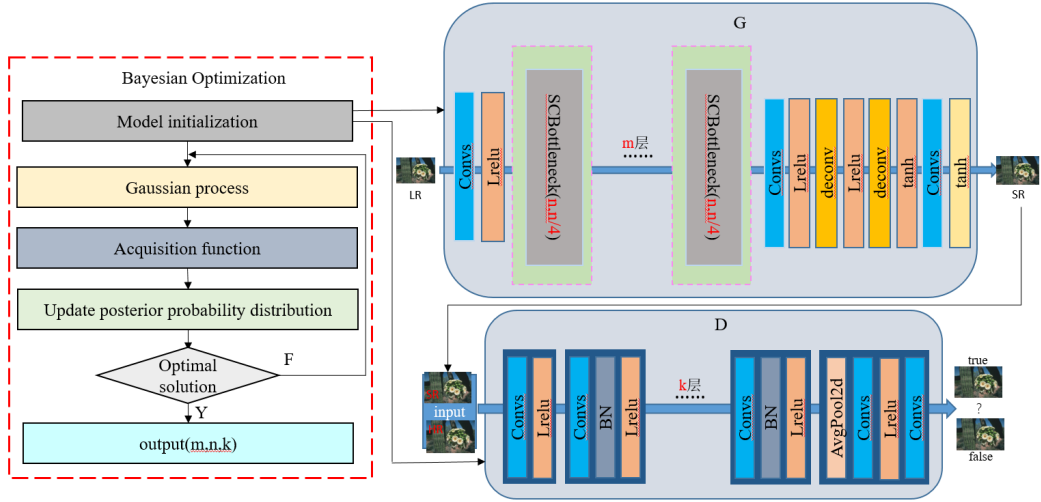}
\caption{The architecture of GAN model with Bayesian optimization}
\label{fig:2}
\end{figure}

According to the four aspects that need to be considered in the use of Bayesian optimization in deep learning, the following specifically introduces the design of super-resolution image reconstruction model by Bayesian optimization method in this paper. The specific network model is shown in figure \ref{fig:2}.

In this paper, we use MSE [21] as the optimization objective function. There are two main reasons for selecting MSE. On the one hand, traditional super-resolution image reconstruction tasks usually use MSE as the loss function. However, the optimization objective function is different from the loss objective function of the model. The optimization objective function is the value that the model expects to be maximized in the Bayesian optimization process (you can choose to minimize, and our method uses the maximize method). Therefore, MSE as the optimization objective function does not conflict with the loss function in super-resolution image reconstruction model. on the other hand, as shown in formula PSNR, in the calculation of the super-resolution image evaluation index PSNR, it is necessary to calculate the MSE value of the two images firstly. In this task, we hope to obtain a set of optimal hyperparameters of the generator, so that our network can reconstruct super-resolution images that are indistinguishable by human eyes. The generative adversarial network is more sensitive to the color of the image, and the reconstructed super-resolution image has good results under the SSIM evaluation index. This chapter is mainly designed to improve the PSNR value. Therefore, MSE is relatively reasonable as an optimization objective function. The optimization objective function formula is:

\begin{center}
$f(I) = \frac{1}{n}\sum\limits_{i = 1}^n {{{({I^{HR}} - {I^{SR}})}^2}} $
\end{center}

An evaluation function $f(I)$ is obtained by above formula. The historical parameter value is replaced by one parameter $\theta $, expressed as $\theta  = \{ ({m_1},{n_1}),({m_2},{n_2}), \cdots ({m_t},{n_t})\} $, and the next set of parameter values is predicted to be $({m_{t + 1}},{n_{t + 1}})$. The model is expressed as $f(I) = (I,\theta )$. When data is continuously input into the model, a set of evaluation function values are obtained from this, expressed as:

\begin{center}
$D = \{ ({I_1},f({I_1})),({I_2},f({I_2})), \cdots ({I_t},f({I_t}))\} $
\end{center}

Based on the above-known sample points, make a prediction to the next sample point. On the basis of these known sample points, the probability distribution of the evaluation function is obtained through the Gaussian process, and the objective function is mapped through the kernel function to make it to the Gaussian distribution. The Gaussian process is expressed as:

\begin{center}
$f(I) \sim GP(\mu ,k(I,I'))$
\end{center}

Where the mean of Gaussian distribution is $\mu $, $k(I,I')$ is the kernel function. After kernel function mapping, the posterior distribution of the new evaluation point is obtained. At the meantime, all the distributions are the Gaussian distribution. The predicted value is expressed as ${y_i} = f({I_i}) + {\varepsilon _i}$. ${\varepsilon _i}$ indicates that variance $\sigma _0^2$ is added to Gaussian noise, and its value is close to 0. The formula $y = {({y_1},{y_1} \cdots {y_t})^N}$ calculates the prior covariance vector value between $y$ and $f(I)$, expressed as $k(f(I),y)$. After calculation, the Gaussian distribution obtained by Gaussian kernel mapping is expressed as follows:

\begin{center}
$p(f({I^ * })|y) = N(f({I^ * })|\mu ({I^ * }),\sigma ({I^ * }))$
\end{center}

Where $\mu ({I^ * })$ is the mean and $\sigma ({I^ * })$ is the variance, both of which are calculated by the covariance.

After the above process, a set of values can be expressed as ${D_1} = (f({I_i}),{y_i})$, so as to predict the next evaluation point. The selection of the next evaluation point is determined by the acquisition function. The acquisition function construction depends on the optimization objective function. A classic evaluation method is:

\begin{center}
$p(f(I) \ge \nu ) = \Phi (\frac{{\mu (I) - \nu }}{{\sigma (I)}})$
\end{center}

Among them, $\nu $ represents the parameter ${y_i}$ that maximizes the predicted value and $\Phi $ is a standard normal distribution. $\mu (I)$ and $\sigma (I)$ are the mean and variance of each evaluation point in the objective function. The mean represents the expected maximum value, and the variance represents the exploration ability of the model.

In this paper we adopt UCB (Upper Confidence Bound) to balance the relationship between mean and variance, and predicting the next evaluation point. Xiang\cite{Xiang2020A} et al has been proved that the UCB method is superior to the EI(Expected Improvement) method in frame construction. UCB formula is as follows:

\begin{center}
$ucb = \mu (I) + \lambda \sigma (I)$
\end{center}

Among them, $\lambda $ is a hyperparameter, which is used to adjust the mean and variance, and the setting value is 1.

The specific algorithm of the method in this paper shows in algorithm ~\ref{algorithm:1}:
\begin{algorithm}[t]
	\caption{Generative Adversarial Networks with Bayesian Optimization}
	\label{algorithm:1}
	\LinesNumbered
	\KwIn{image datasets}
    initialize training iterations(number), batch size(step)\\
    initialize the boundary of hyperparameters(m, n, k)\\
	\For {i in iterations}
     { \For {number in training iterations}
          {
          \For {step in steps}
            {
              ${\nabla _{{\theta _d}}}\frac{1}{m}\sum\limits_{i = 1}^m {\left[ {\log D({x^{(i)}}) + \log (1 - D(G({z^i})))} \right]} $
             }
           ${\nabla _{{\theta _g}}}\frac{1}{m}\sum\limits_{i = 1}^m {\left[ {\log (1 - D(G({z^i})))} \right]} $
          }
	   $y \leftarrow f(I,\theta )$\\
       ${\theta _{t + 1}} \leftarrow {\theta _{1:t}}$\\
       $re = \arg \max {y_i}$
       return $re$
     }
   \KwOut{Objective function value}
\end{algorithm}

\section{Experiment}

This chapter first introduces the experimental conditions and the training details, then it introduces and analyzes the experimental results, finally it shows the effect of applying the above optimization results to super-resolution image reconstruction tasks.

\subsection{Experimental conditions}

The experiment was trained on the Windows 10 operating system and NVIDIA 2080Ti server. The CUDA version is 10.2. All source programs are written in python language and implemented using the pytorch framework. The pytorch version is 1.6. The batch size is 64. The upper and lower bounds of hyperparameters are [2, 11], [64, 256] and [2,10] respectively. The method based on Bayesian optimization uses a part of the VOC2012 \cite{Ledig2016Photo} datasets, and using the test set in the VOC2012 as the training set. Since the batch size is 64, a certain number of images were randomly selected from the original training set to made the total number of 64 multiple. There were eight images been selected from the original training set as the verification set, which were used in the calculation of the objective function in the optimization algorithm.

\subsection{Evaluation index}

We evaluated the reconstructed super-resolution image under the evaluation indicators PSNR and SSIM \cite{Zhou2004Image}. PSNR compared the corresponding pixels between super-resoluton image and high-resolution image. SSIM compared the similarity between super-resoluton image and high-resolution image.

The calculation formula of MSE is:

\begin{center}
$MSE = \frac{1}{n}\sum\limits_{i = 1}^n {{{({I^{HR}} - {I^{SR}})}^2}} $
\end{center}

\begin{center}
$PSNR = 10 \cdot {\log _{10}}\left( {{1 \mathord{\left/
 {\vphantom {1 {MSE}}} \right.
 \kern-\nulldelimiterspace} {MSE}}} \right)$
\end{center}
Where high-resolution image is expressed as ${I^{HR}}$, and super-resolution image is expressed as ${I^{SR}}$.

The calculation formula of SSIM is:

\begin{center}
$SSIM = \frac{{\left( {2{\mu _{{I^{SR}}}}{\mu _{{I^{HR}}}} + {C_1}} \right)\left( {2{\sigma _{{I^{SR}}{I^{HR}}}} + {C_2}} \right)}}{{\left( {\mu _{{I^{SR}}}^2 + \mu _{{I^{HR}}}^2 + C1} \right)\left( {\sigma _{{I^{SR}}}^2 + \sigma _{{I^{HR}}}^2 + C1} \right)}}$
\end{center}
Where ${\mu _{{I^{SR}}}}$ is the mean of ${I^{SR}}$, ${\mu _{{I^{HR}}}}$ is the mean of ${I^{HR}}$, ${\sigma _{{I^{SR}}{I^{HR}}}}$ is the covariance of ${I^{SR}}$ and ${I^{HR}}$, and $\sigma _{{I^{SR}}}^2$ and $\sigma _{{I^{HR}}}^2$ is the variance of ${I^{SR}}$ and ${I^{HR}}$.

The greater the value of the above evaluation method, the higher the quality of the reconstructed image, and the more difficult to distinguish for the human eye.

\subsection{Experimental results and analysis}

This section mainly focuses on the display and analysis of the results of Bayesian optimization and the results of the super-resolution image reconstruction of the generative adversarial network. First, a brief description of the relevant algorithms:

\begin{itemize}
\item COBYLA algorithm \cite{Powell1994A}: it is a linear approximation optimization algorithm that constrains the region bound. Calculate the upper and lower bounds based on the simplex, and then determine the region bound. when the objective function is approximated by linear interpolation, the optimal solution constructed should be within the region bound. This algorithm has a better effect on integer boundary constraints, but it will make mistakes on decimal boundary constraints.
\item PSO algorithm \cite{Kennedy2002Particle}: this algorithm has two attributes: position and speed, which respectively describe the direction of exploration and the speed of movement when the individual approaches the target value. Individuals in the group need to exchange information to select the individual who currently reaches the best target value. When the algorithm has a large scale, as the individual or dimension increases, the amount of calculation also increases.
\item Bayesian optimization algorithm \cite{Pelikan1999BOA}: it is an optimization algorithm based on the Bayesian probability model. Unlike the above method, this method will refer to the previous evaluation results when looking for the next set of solutions. Bayesian optimization consists of four parts: recording historical results is convenient for trying the next set of solution references; defining upper and lower bounds of hyper-parameters; adopting the loss on the verification set as the objective function; constructing the collection function to select the next evaluation point.
\end{itemize}

\begin{figure}[t]
\centering
\includegraphics[scale=0.6]{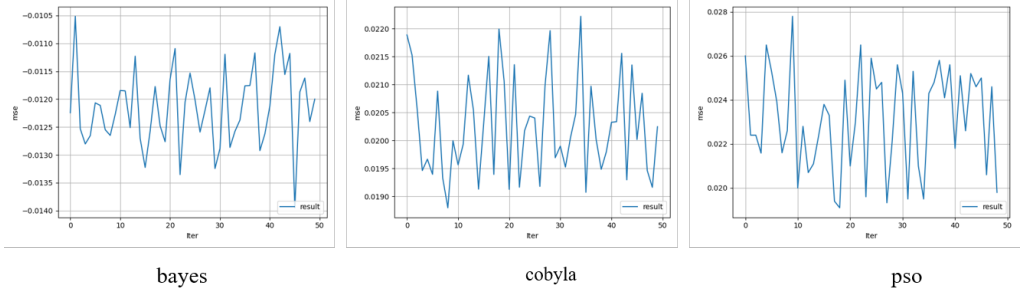}
\caption{The curve of $BOP_G + BOP_D$ objection function}
\label{fig:3}
\end{figure}

\begin{figure}[t]
\centering
\includegraphics[scale=0.6]{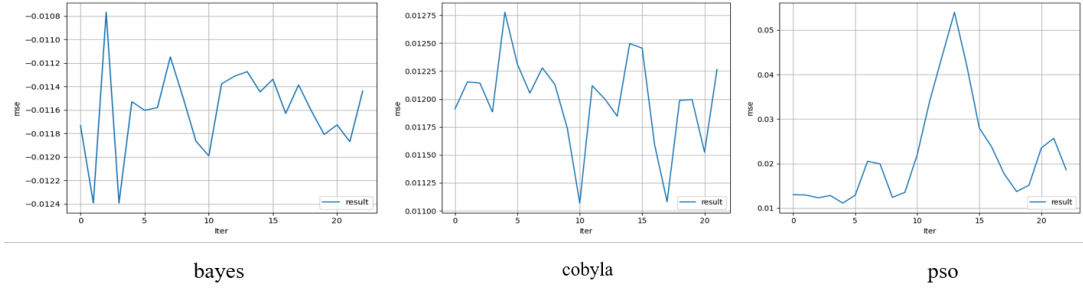}
\caption{The curve of $BOP_G$ objection function}
\label{fig:4}
\end{figure}

In order to illustrate the superiority of the Bayesian optimization algorithm, we compared the experimental results of two generative adversarial network construction methods. The first is the search result of the generator and the discriminator parameter by the optimization algorithm ($BOP_G + BOP_D$); the second is the search for the generator parameter by the optimization algorithm ($BOP_G$). Since the generator in this article is more complicated than the discriminator, the generator is searched separately. The experimental results of the above three optimization algorithms for the optimization of the two network structures are shown in figure \ref{fig:3} and figure \ref{fig:4}.

In the figure \ref{fig:3} and figure \ref{fig:4}, the horizontal axis is the number of iterations, and the vertical axis is the optimized objective function value. Bayesian optimization searches for a set of hyperparameters that can maximize the objective function, and COBYLA and PSO optimization methods search for a set of hyperparameters that minimize the objective function. The $BOP_G+BOP_D$ optimization results can be obtained from figure \ref{fig:3}. The Bayesian optimization algorithm maximizes the objective function at the second iteration, the COBYLA algorithm minimizes the objective function at the 9th iteration, and the PSO algorithm minimizes the objective function at the 19th iteration. The $BOP_G$ optimization results can be obtained from figure \ref{fig:4}. The Bayesian optimization algorithm maximizes the objective function at the 3rd iteration, the COBYLA algorithm minimizes the objective function at the 11th iteration, and the PSO algorithm minimizes the objective function at the 5th iteration. It can be seen that Bayesian optimization can search for a set of hyperparameters that meet the conditions in the least number of iterations, which verifies the effect of the Bayesian optimization algorithm.

Through the above-mentioned optimization algorithm curve, we obtain the optimal hyperparameter combinations of the two network models $BOP_G+BOP_D$ and $BOP_G$ under different algorithms. The specific hyperparameters of different algorithms are shown in table \ref{tab:1} and \ref{tab:2}.

\begin{table*}[htb]
\centering
\begin{center}
\caption{$BOP_G+BOP_D$ model hyperparameter result} \label{tab:1}
\begin{tabular}{cccc}
\hline
\toprule
hyperparameter&Bayesian optimization&COBYLA&PSO\\
\hline
Layers(G) & 3 & 2 & 2 \\
Neurons(G) & 140 & 128 & 132 \\
Layers(D) & 3 & 2 & 2 \\
\bottomrule
\end{tabular}
\end{center}
\end{table*}

\begin{table*}[htb]
\centering
\begin{center}
\caption{$BOP_G$ model hyperparameter result} \label{tab:2}
\begin{tabular}{cccc}
\hline
\toprule
hyperparameter&Bayesian optimization&COBYLA&PSO\\
\hline
Layers(G) & 3 & 2 & 2 \\
Neurons(G) & 184 & 124 & 160 \\
\bottomrule
\end{tabular}
\end{center}
\end{table*}

According to the network parameters obtained by the above-mentioned different optimization algorithms, a super-resolution image reconstruction generative adversarial network model ($BOP_G+BOP_D$) is constructed. In the case of scale 4, we use all the data of the VOC2012 to train the generative adversarial network to obtain a super-resolution image reconstruction model.
The model performs on the above three algorithms in the set5, set14 and BSD100 test sets in turn. The comparison results under each evaluation index are shown in table \ref{tab:3}. In addition, this paper also verified the experimental results of the $BOP_G$ model on different test sets, and the comparison results of the reconstructed super-resolution images in each evaluation index are shown in the table \ref{tab:4}.

\begin{table*}[htb]
\centering
\begin{center}
\caption{$BOP_G+BOP_D$ model results under different test sets} \label{tab:3}
\begin{tabular}{ccccc}
\hline
\toprule
Datasets&Methon&Bayesian optimization&COBYLA&PSO\\
\hline
\multirow{2}*{set5} & PSNR & 29.89 & 29.77 & 29.49 \\
 ~& SSIM & 0.9288 & 0.9277 & 0.9240 \\
 \hline
\multirow{2}*{set14} & PSNR & 26.59 & 26.46 & 26.33 \\
 ~& SSIM & 0.8531 & 0.8532 & 0.8505 \\
 \hline
\multirow{2}*{BSD100} & PSNR & 25.97 & 25.95 & 25.89 \\
 ~& SSIM & 0.8107 & 0.8115 & 0.8062 \\
\bottomrule
\end{tabular}
\end{center}
\end{table*}

\begin{table*}[htb]
\centering
\begin{center}
\caption{$BOP_G$ model results under different test sets} \label{tab:4}
\begin{tabular}{ccccc}
\hline
\toprule
Datasets&Methon&Bayesian optimization&COBYLA&PSO\\
\hline
\multirow{2}*{set5} & PSNR & 29.00 & 27.40 & 28.85 \\
 ~& SSIM & 0.9131 & 0.9139 & 0.9143 \\
 \hline
\multirow{2}*{set14} & PSNR & 26.11 & 25.14 & 26.02 \\
 ~& SSIM & 0.8419 & 0.8400 & 0.8424 \\
 \hline
\multirow{2}*{BSD100} & PSNR & 25.62 & 24.88 & 25.60 \\
 ~& SSIM & 0.8013 & 0.8002 & 0.8015 \\
\bottomrule
\end{tabular}
\end{center}
\end{table*}

This paper mainly designs and optimizes the objective function for the PSNR value, which promotes the model to have better results under the evaluation index PSNR. From the table \ref{tab:3} and table \ref{tab:4}, we have compared the PSNR values of two different network models ($BOP_G+BOP_D$ and $BOP_G$ )to reconstruct super-resolution images on different test data sets. The experimental results of the two network models are as follows. The result of $BOP_G$ is: the result of COBYLA algorithm is the worst. The result of PSO algorithm is relatively close to that of Bayesian optimization algorithm, but the result of Bayesian optimization algorithm is slightly better. The result of $BOP_G+BOP_D$ is: the result of PSO optimization algorithm is poor, the result of COBYLA optimization algorithm is relatively close to that of Bayesian optimization algorithm, but the result of Bayesian optimization algorithm is the best.
Under the evaluation index SSIM, the results of the three optimization algorithms are relatively close. Because the method in this paper is mainly optimized for the MSE objective function, and the result of reconstructing the image under the PSNR evaluation index is more obvious, which achieves the purpose of this paper.

\begin{figure}[t]
\centering
\includegraphics[scale=0.6]{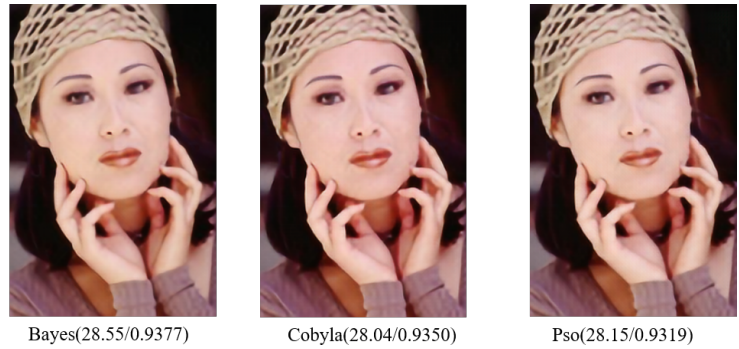}
\caption{Visualization of the super-resolution image of $BOP_G+BOP_D$}
\label{fig:5}
\end{figure}

This paper not only compares the results of each evaluation index, but also gives the visualization results of the reconstructed image. The super-resolution image reconstruction results of the three optimization algorithms under the set5 and set14 are shown in figure \ref{fig:5}. It can be seen from the figure that Bayesian optimization has better visual effects compared with the other two algorithms, regardless of the $BOP_G+BOP_D$ and the $BOP_G$.

\begin{figure}[t]
\centering
\includegraphics[scale=0.6]{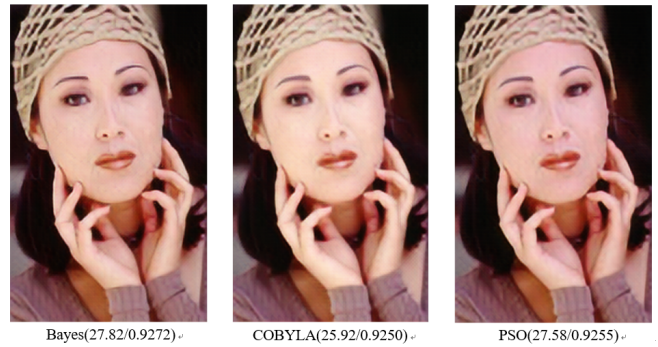}
\caption{Visualization of the super-resolution image of $BOP_G$}
\label{fig:6}
\end{figure}

In summary, since the training of the generative adversarial network is a max-min game process, the experimental results of the $BOP_G+BOP_D$ model under the three optimization algorithms are better than the $BOP_G$ results, and the Bayesian optimization algorithm has better results in $BOP_G+BOP_D$ and $BOP_G$. The search speed of the PSO algorithm gradually increases when the model scale becomes larger. The PSO algorithm needs to build a topological structure to assist in the transmission of information in the process of searching for the optimal solution. $BOP_G+BOP_D$ adds discriminator parameters, and a lot of calculations are added in the process of information transmission. Therefore the longest search time. The experimental results of COBYLA optimized $BOP_G+BOP_D$ model is closer to the Bayesian optimization algorithm, but there may be dimensional problems. The Bayesian optimization algorithm is a black-box optimization algorithm. There is no complicated topological structure, the optimal value can be searched earliest, and two different reconstructed super-resolution image models have the best effect under the evaluation index.

\section{Conclusion}

We presented methods for performing Bayesian optimization for hyperparameter selection of generative adversarial networks. We introduced a fully Bayesian treatment for UCB, and algorithms for dealing with optimizing generative adversarial network training. The effectiveness of our approaches were demonstrated on the speed of search and reconstruction image quality in different algorithms. The results of Bayesian optimization have found the better hyperparameters which significantly faster than the approaches used by the artificial. It also shows that  Bayesian optimization has a better result than COBYLA and PSO algorithm. However, it is a challenge for us to solve the problem that the quality of the super-resolution image reconstructed by the Bayesian optimization network is worse than the super-resolution image reconstructed by the artificial tuning parameter construction network.

\section*{Acknowledgment}
This work are sponsored by the key project of the National Natural Science Foundation of China (61602421), China Postdoctoral Foundation (2016M600584), Aviation Science Foundation (201828X4001) and Zhengzhou University.

\bibliographystyle{IEEEtran}
\bibliography{reference}

\begin{thebibliography}{10}
\providecommand{\url}[1]{#1}
\csname url@samestyle\endcsname
\providecommand{\newblock}{\relax}
\providecommand{\bibinfo}[2]{#2}
\providecommand{\BIBentrySTDinterwordspacing}{\spaceskip=0pt\relax}
\providecommand{\BIBentryALTinterwordstretchfactor}{4}
\providecommand{\BIBentryALTinterwordspacing}{\spaceskip=\fontdimen2\font plus
\BIBentryALTinterwordstretchfactor\fontdimen3\font minus
  \fontdimen4\font\relax}
\providecommand{\BIBforeignlanguage}[2]{{%
\expandafter\ifx\csname l@#1\endcsname\relax
\typeout{** WARNING: IEEEtran.bst: No hyphenation pattern has been}%
\typeout{** loaded for the language `#1'. Using the pattern for}%
\typeout{** the default language instead.}%
\else
\language=\csname l@#1\endcsname
\fi
#2}}
\providecommand{\BIBdecl}{\relax}
\BIBdecl

\bibitem{Lally2020Unbalanced}
P.~J. Lally, P.~M. Matthews, and N.~K. Bangerter, ``Unbalanced ssfp for
  super-resolution in mri,'' \emph{Magnetic Resonance in Medicine}, vol.~85,
  no.~5, 2020.

\bibitem{Bai2018Finding}
Y.~Bai, Y.~Zhang, M.~Ding, and B.~Ghanem, ``Finding tiny faces in the wild with
  generative adversarial network,'' in \emph{2018 IEEE/CVF Conference on
  Computer Vision and Pattern Recognition}, 2018.

\bibitem{Florea2019Weighted}
A.~C. Florea and R.~Andonie, ``Weighted random search for hyperparameter
  optimization,'' \emph{International Journal of Computers, Communications and
  Control (IJCCC)}, vol.~14, no.~2, pp. 154--169, 2019.

\bibitem{Pelikan1999BOA}
M.~Pelikan, ``Boa: The bayesian optimization algorithm,'' in \emph{Proc Genetic
  and Evolutionary Computation Conference}, 1999.

\bibitem{Goodfellow2014Generative}
I.~Goodfellow, J.~Pouget-Abadie, M.~Mirza, B.~Xu, D.~Warde-Farley, S.~Ozair,
  A.~ourville, and Y.~Bengio, ``Generative adversarial networks,''
  \emph{Advances in Neural Information Processing Systems}, vol.~3, pp.
  2672--2680, 2014.

\bibitem{Liu2018Image}
G.~Liu, F.~A. Reda, K.~J. Shih, T.~C. Wang, A.~Tao, and B.~Catanzaro, ``Image
  inpainting for irregular holes using partial convolutions,'' in
  \emph{European Conference on Computer Vision}, 2018.

\bibitem{Sabini2018Painting}
M.~Sabini and G.~Rusak, ``Painting outside the box: Image outpainting with
  gans,'' 2018.

\bibitem{Wang2017High}
T.~C. Wang, M.~Y. Liu, J.~Y. Zhu, A.~Tao, J.~Kautz, and B.~Catanzaro,
  ``High-resolution image synthesis and semantic manipulation with conditional
  gans,'' 2017.

\bibitem{Chen2021Cascading}
J.~Chen, Y.~Zhang, X.~Hu, and Y.~C. Chen, ``Cascading residual-residual
  attention generative adversarial network for image super resolution,''
  \emph{Soft Computing}, no.~1, 2021.

\bibitem{Chudasama2021Computationally}
V.~Chudasama and K.~Upla, ``Computationally efficient progressive approach for
  single-image super-resolution using generative adversarial network,''
  \emph{Journal of Electronic Imaging}, vol.~30, no.~2, 2021.

\bibitem{Chudasama2021RSRGAN}
------, ``Rsrgan: computationally efficient real-world single image
  super-resolution using generative adversarial network,'' \emph{Machine Vision
  and Applications}, vol.~32, no.~2, pp. 1--18, 2021.

\bibitem{Bergstra2011Algorithms}
J.~Bergstra, R.~Bardenet, B.~Kegl, and Y.~Bengio, ``Algorithms for
  hyper-parameter optimization,'' in \emph{Advances in Neural Information
  Processing Systems}, 2011.

\bibitem{Audet2017Nelder}
C.~Audet and W.~Hare, ``Nelder-mead,'' 2017.

\bibitem{Hz2020Orthogonal}
A.~Hz, C.~Aahb, M.~W. D, E.~Lz, A.~Hc, and F.~Cl, ``Orthogonal nelder-mead moth
  flame method for parameters identification of photovoltaic modules,''
  \emph{Energy Conversion and Management}, vol. 211.

\bibitem{Niegodajew2020Power}
P.~Niegodajew, M.~Marek, W.~Elsner, and U.~Kowalczyk, ``Power plant
  optimisation-effective use of the nelder-mead approach,'' \emph{Processes},
  vol.~8, no.~3, p. 357, 2020.

\bibitem{Powell1994A}
M.~Powell, ``A direct search optimization method that models the objective and
  constraint functions by linear interpolation,'' \emph{Advances in
  Optimization and Numerical Analysis}, vol. 275, pp. 51--67, 1994.

\bibitem{Kramer2016Scikit}
O.~Kramer, \emph{Scikit-Learn}.\hskip 1em plus 0.5em minus 0.4em\relax Machine
  Learning for Evolution Strategies, 2016.

\bibitem{Kennedy2002Particle}
J.~Kennedy and R.~Eberhart, ``Particle swarm optimization,'' in
  \emph{Icnn95-international Conference on Neural Networks}, 2002.

\bibitem{Gong2020Research}
D.~Gong, ``Research on hyperparameter evolutionary tuning of deep neural
  network mode and its application,'' in \emph{Master's thesis}, 2020.

\bibitem{Liu2021Industrial}
H.~Liu and Z.~Zhou, ``Industrial control network intrusion detection based on
  hyperparameter automatic optimization,'' \emph{Information and control}, no.
  1-8, 2021.

\bibitem{Xiang2020A}
Z.~Xiang, C.~Xiang, T.~Li, and Y.~Guo, ``A self-adapting hierarchical actions
  and structures joint optimization framework for automatic design of robotic
  and animation skeletons,'' \emph{Soft Computing}, no.~3, 2020.

\bibitem{Xu2020A}
Y.~Xu, X.~Yang, Y.~Gan, W.~Zhou, H.~Cheng, and X.~He, ``A music generation
  model based on generative adversarial networks with bayesian optimization,''
  in \emph{Chinese Intelligent Systems Conference}, 2020.

\bibitem{Ledig2016Photo}
C.~Ledig, L.~Theis, F.~Huszar, J.~Caballero, A.~Cunningham, A.~Acosta,
  A.~Aitken, A.~Tejani, J.~Totz, and Z.~Wang, ``Photo-realistic single image
  super-resolution using a generative adversarial network,'' \emph{IEEE
  Computer Society}, 2016.

\bibitem{Zhou2004Image}
W.~Zhou, A.~C. Bovik, H.~R. Sheikh, and E.~P. Simoncelli, ``Image quality
  assessment: from error visibility to structural similarity,'' \emph{IEEE
  Trans Image Process}, vol.~13, no.~4, 2004.

\end{thebibliography}
\ifCLASSOPTIONcaptionsoff
  \newpage
\fi

\end{document}